\documentclass[pdflatex,sn-mathphys-num]{sn-jnl}

\usepackage{graphicx}%
\usepackage{multirow}%
\usepackage{amsmath,amssymb,amsfonts}%
\usepackage{amsthm}%
\usepackage{mathrsfs}%
\usepackage[title]{appendix}%
\usepackage{xcolor}%
\usepackage{textcomp}%
\usepackage{manyfoot}%
\usepackage{booktabs}%
\usepackage{algorithm}%
\usepackage{algorithmicx}%
\usepackage{algpseudocode}%
\usepackage{listings}%

\theoremstyle{thmstyleone}%
\theoremstyle{thmstyletwo}%

\theoremstyle{thmstylethree}%

\begin{document}

\title[Article Title]{Detecting AI-Generated Images via CLIP}


\author*[1]{\fnm{Alexander} \sur{Moskowitz}}\email{alex.m@visimo.ai}

\author[1]{\fnm{Tyler} \sur{Gaona}}

\author[2]{\fnm{Jacob} \sur{Peterson}}

\affil[1]{ \orgname{VISIMO}, \orgaddress{\street{ 520 East Main Street}, \city{Carnegie}, \postcode{15106}, \state{PA}, \country{USA}}}

\affil[2]{\orgname{Haemonetics Corporation}, \orgaddress{\street{125 Summer Street}, \city{Boston}, \postcode{02110}, \state{MA}, \country{USA}}}


\abstract{As AI-generated image (AIGI) methods become more powerful and accessible, it has become a critical task to determine if an image is real or AI-generated. Because AIGI lack the signatures of photographs and have their own unique patterns, new models are needed to determine if an image is AI-generated. In this paper, we investigate the ability of the Contrastive Language–Image Pre-training (CLIP) architecture, pre-trained on massive internet-scale data sets, to perform this differentiation. We fine-tune CLIP on real images and AIGI from several generative models, enabling CLIP to determine if an image is AI-generated and, if so, determine what generation method was used to create it. We show that the fine-tuned CLIP architecture is able to differentiate AIGI as well or better than models whose architecture is specifically designed to detect AIGI. Our method will significantly increase access to AIGI-detecting tools and reduce the negative effects of AIGI on society, as our CLIP fine-tuning procedures require no architecture changes from publicly available model repositories and consume significantly less GPU resources than other AIGI detection models.}




\maketitle

\section{Introduction}\label{section_intro}
The past few years have seen a meteoric increase in the quality of AI-generated images (AIGI).
The pairing of internet-scale datasets with new models based on techniques such as Diffusion \cite{stablediffusion}; Diffusion-based models (see Table \ref{tab:generators}) are overtaking the previous state-of-the-art Generative Adversarial Network (GAN) models, which are difficult to train due to mode collapse, lack of convergence, vanishing gradients, and overall training instability \cite{GAN}. Large-scale diffusion models can create a wide variety of images from a single model, even allowing the user to specify the artistic style of the output image.

The impact of AIGI has been amplified by the relative accessibility of AIGI models to the public. With cloud-hosted, web-based natural language models accessed via popular platforms such as Discord, no programming knowledge or expensive GPU-enabled hardware is required to use generative models. Many AIGI model hosts offer free trial accounts, and paid versions cost only a handful of dollars per month \cite{pricing-1,pricing-2}. The combination of speed, quality, and availability have led to a deluge of AIGI content posted and shared across the internet. 

The scale and quality of AI-generated content exacerbates old problems and introduces entirely new ones. Conflicts often arise because AIGI are now difficult to distinguish from real images, and affected parties may only discover an image is synthetic after damage has been done \cite{art-contest}. Copyright of AI-generated content is thorny, as traditionally only humans are allowed to hold copyright \cite{nyt-art}. It is not clear if copyright is even applicable in the situation, as trained AI models do not store content after training or copy content during generation \cite{lawsuit}. While edge cases have always existed in copyright law such as copyright of animal-created works \cite{monkey}, the sheer number of training images used and the increase in AIGI  use now requires clear solutions to these issues.

A further problem is the ability of AIGI to generate enormous amounts of harmful content, either offensive or misleading. AIGI tools allow easy, targeted generation or modification of images on a massive scale with no prerequisites for artistic or computational skill. This creates potential societal-level impacts; AIGI can mislead the public via social media or advertising. The ability to create multiple consistent fake images from varying perspectives may lend a weight of realism to misinformation campaigns that a single fake image could not. AIGI are also a concern for organizations that rely on publicly available social media data, including intelligence agencies and military branches \cite{social-media}. 

Despite efforts by generative AI developers to limit the generation of harmful content, such rules often have loopholes \cite{jailbreaking}. The datasets scraped from the internet and used to train AIGI models can also be poisoned to either break the model \cite{poison} or cause it to create harmful content \cite{tay}. In fact, AIGI models might poison themselves by ingesting their own generated content as training data \cite{self-poison}.

Many of these problems can be mitigated or eliminated outright if widepspread, reliable AIGI detection tools are available. Automatically discarding AIGI fights misinformation and solves the data poisoning problems, but this must be done on a scale greater than human moderators can hope to accomplish. As the legal boundary between AI-generated and human-generated art blurs, determining whether an image was wholly produced by an algorithm becomes important. AIGI detection models also pose an interesting theoretical challenge, as they contain crucial differences from photographic image manipulation detection models, which rely on features such as camera noise \cite{MVSS, camera-type} or compression artifacts \cite{CAT-NET} that are not present in AIGI. 

Accordingly, specialized ML models have been developed to identify AIGI images. However, in this paper, we propose an entirely novel technique. We start with a generalized model backbone pretrained on internet-scale datasets, and fine-tune the model and head layers to perform AIGI detection. We use the pretrained CLIP model \cite{CLIP} as our backbone, and fine-tune the model using images generated by several different generative methods. Despite our model lacking any AIGI-specific architecture, the power of the internet-scale pretraining allows us to outperform specialized AIGI detection models.

This paper is structured as follows. In Section \ref{section_related_work}, we discuss other approaches to the detection of AIGI. In Section \ref{section_methods}, we introduce the CLIP model, summarize our training dataset, and describe our training procedures. In Section \ref{section_results}, we report on the performance of our model and compare our results to previous approaches. Finally, in section \ref{section_conclusion}, we summarize our work and discuss its implications. 

\section{Related Work}\label{section_related_work}
There exist several recent deep learning approaches for AIGI detection. The authors of \cite{cnndet} cite a basic method for detection of AIGI generated by a single model: train a binary classifier on a set of real images and fake images generated by the model. Prior to the authors' work, this simple approach suffered from failure to generalize to new data and failure to generalize to fake images generated by different techniques. The authors showed that a binary classifier trained on a large number of fake images generated by a single CNN model was able to generalize to fake images generated by a wide variety of CNN models. The authors claim that data augmentation in the form of image post-processing and training set diversity were critical to the success of their model. Henceforth, we refer to this model as \textit{CNNDet}. 

The authors of \cite{diffdet} sought to investigate if approaches such as \cite{cnndet} and \cite{gragnaniello} would also generalize to the detection of fake images generated by diffusion models. They find a significant performance decrease when the pretrained models of \cite{cnndet} and \cite{gragnaniello} are applied to diffusion generated images, but are able to recover the performance by finetuning the aforementioned models on diffusion generated images.

In \cite{dire}, the authors also train a binary classifier to detect diffusion generated images but provide a different input to the classifier. Their hypothesis is that images produced by diffusion processes can be reconstructed more accurately by a pretrained diffusion model compared to real images. Given a diffusion model that provides mappings \(I\) and \(R\) which respectively invert an image into noise and reconstruct an image from noise, an input image \(\mathbf{x}\) is inverted and reconstructed into \(\mathbf{x}^\prime = R(I(\mathbf{x}))\). The Diffusion Reconstruction Error (DIRE) of the input image is then defined as 
\[\text{DIRE}(\mathbf{x}) = |\mathbf{x} - R(I(\mathbf{x}))|\]
where \(| \cdot |\) denotes the absolute value applied pixelwise. Restated in terms of DIRE, the authors' hypothesis is that diffusion generated images have DIRE values closer to zero than real images. The authors report better accuracy and average precision at detecting diffusion generated images than the method of \cite{cnndet} retrained to detect images generated by the ADM model of \cite{ADM}. The authors do not report how their method performs on fake images generated by non-diffusion models, such as CNN-based GANs. 
\section{Methods}\label{section_methods}
\subsection{CLIP Model}\label{subsection_clip}

The CLIP model \cite{CLIP} learns image and language concepts by relating image embeddings to matching text embeddings. It is trained in a ``contrastive" manner where pairings are used in both a positive and negative manner; the model learns to maximize the cosine similarity between matching image and text embeddings, and learns to minimize the cosine similarity between non-matching image and text embeddings. Both positive and negative pairs are weighted equally when applied to the model's Cross Entropy loss function. Because of this, the CLIP architecture is naturally suited to image classification tasks - it simply calculates the cosine similarity between the image and all label categories and chooses the one with the highest cosine similarity. The CLIP model trains much more efficiently than similar vision-language models, due to features such as a Bag-of-Words (BoW) tokenization scheme, linear projections for embeddings, and treatment of temperature as a model parameter instead of a hyperparameter. This allows CLIP to take full advantage of its training dataset's size and breadth. The pretrained CLIP model is trained on roughly 400 million image/text pairings that has been specifically curated to cover a wide range of topics. The individual embedding models use either a ResNet or Vision Transformer backbone for image embedding, and a Transformer architecture for text embedding. 

Our motivation for using CLIP for AIGI detection is its remarkable ability to adapt to new image processing tasks, showing promise with many zero-shot challenges \cite{CLIP}. CLIP excels at adapting to shifting input domain, with its largest improvement over other models happening when the image style is changed. This is exactly the behavior we desire from a AIGI detection model, as we seek to identify the patterns arising from AIGI independent of its content. In contrast, the CLIP model suffers during content-specific tasks, struggling to count objects in an image and failing to retrieve information about  specific knowledge domains such as medical images. As our classification task is content-agnostic and our dataset covers a wide variety of contexts, this area of poor performance is less of a concern.

\subsection{Datasets}\label{subsection_datasets}
It is insufficient for a model to proficiently detect AIGI coming from a single generation source, due to the breadth and rapid improvement in generative models. Accordingly, to train and test the CLIP models in our approach, we acquired images produced by a number of generative models, including both diffusion and GAN models. The final dataset drew AIGI from \cite{towards}, while real images were drawn from from  the bedroom subset of \cite{lsun}. The generation methods used in the training set are shown in Table \ref{tab:generators}. Overall, we used 1000 images from each generation method in the training dataset, and the same amount in the testing dataset; for the real images, images were again  taken from \cite{lsun}. 

\subsection{CLIP Fine-Tuning}\label{subsection_fine_tuning}
 Because CLIP operates by choosing a caption from among a set of captions, the base CLIP architecture is already configured to perform a classification task; to label our dataset for CLIP, we assigned each image generation method, as well as the real images, a unique caption. \cite{CLIP} show that prompt engineering can improve CLIP performance on new tasks. Following their style, we begin each label with ``an image of a ", which gives the model the appropriate context to perform classification task about the image itself. The CLIP model uses the BoW tokenization method, and so word order is not important; accordingly, we ensure our labels include 1) if the image is real or fake 2) if the image uses a diffusion model, and 3) the specific name of the generating model. These captions are shown in Table \ref{tab:generators}.
 
 To fine-tune the CLIP model to perform AIGI model differentiation, we started with a pre-trained CLIP model with a ResNet101 image encoder and the default CLIP text encoder, both available via CLIP's published source code repository \cite{clip-repo}. We then trained the CLIP model on the dataset; our final model used the hyperparameters listed in Table \ref{tab:hyper}, where $\beta_1$ and $\beta_2$ refer to the parameters of the Adam optimizer. The CLIP training method consists of feeding image-caption pairs to the image and text encoders, projecting the embeddings to a shared plane, and determining the cosine similarity between the image-caption embeddings in the forward loop.  The model weights are learned using the combined image and text Cross Entropy losses.
\renewcommand{\tabcolsep}{2pt}
 \begin{table*}[htbp]
  \centering
  \footnotesize
  \caption{Generation Methods and Caption Labels}
 \hspace*{-2.0cm}
  \begin{tabular}{|c|c|c|c|c|}
    \hline
    Method & Generator Type & Caption & Source & Abbreviation\\
    \hline
    Ablated Diffusion & Diffusion & ``a fake image from ablated diffusion" & \cite{ADM} & ADM\\
    Probabilistic Denoising Diffusion & Diffusion & ``a fake image from denoising diffusion" & \cite{DDPM}& DDPM\\
    Pseudo Numerical Diffusion & Diffusion & "a fake image from psuedo numerical diffusion" & \cite{pseudo} &PNDM \\
    Improved Probabilistic Denoising Diffusion & Diffusion & ``a fake image from improved denoising diffusion" & \cite{improved} & IDDPM\\
Latent Diffusion & Diffusion & "a fake image from latent diffusion" & \cite{stable}& LDM\\
ProjectedGAN & GAN & "a fake image from original ProjectedGAN" & \cite{projectedGAN} & PjG \\
StyleGAN & GAN & "a fake image from original StyleGan" & \cite{stylegan} & SG \\  
ProGAN & GAN & a fake image from ProGAN & \cite{progan}  & PG\\
Diff-ProjectedGAN & GAN & "a fake image from Diff-ProjectedGAN" & \cite{diffprogan} & DPjG\\
Diff-StyleGAN2 & GAN & "a fake image from Diff-StyleGAN2" & \cite{diffprogan} & DSG \\
Real Image & & "a real image with no alterations" & \cite{lsun}& \\
    \hline
  \end{tabular}
  \label{tab:generators}
\end{table*}
 
 \begin{table}[htbp]
  \centering
  \caption{Final Training Hyperparameters}
  \begin{tabular}{|c|c|}
    \hline
    Hyperparameter & Value \\
    \hline
    Optimizer & Adam \\
    Epochs & 12 \\
    Batch Size & 16 \\
    Learning Rate & $10^{-6}$ \\
    $\beta_{1}$ & 0.9 \\
    $\beta_{2}$ & 0.98 \\
    eps & $10^{-6}$ \\
    Weight Decay & $10^{-4}$\\
    \hline
  \end{tabular}
  \label{tab:hyper}
\end{table}

\section{Results}\label{section_results}
For our final evaluation dataset, we used 1000 images created by each generative method, as well as 1000 real images. 

Tables \ref{tab:acc} - \ref{tab:clip} show the numerical results of our experiments. Table \ref{tab:acc} shows the accuracy  when the CNNDet, DIRE, and CLIP models are tested on  each generation method. Table \ref{tab:count} shows the same data as Table \ref{tab:acc}, but in terms of raw numbers of images. Finally, Table \ref{tab:clip} shows the confusion matrix for the CLIP model as well as CLIP's Precision, Recall, and F1 scores for each generation method. The high values of these specific metrics for real images confirm that the accuracy values for real/AIGI image differentiation implied by Table \ref{tab:acc} are not biased by the number imbalance between all real and all AIGI images. Overall, we see that the fine-tuned CLIP model performs far above CNNDet, and performs near or above DIRE.

 We observe that CLIP has some trouble distinguishing images generated by ADM \cite{ADM} and IDDPM \cite{improved} and images generated by Diff-ProjectedGAN \cite{diffprogan} and ProjectedGAN \cite{projectedGAN}. In each case, the one model is derived from the other or otherwise shares many similarities. For other generation methods, our finetuned CLIP model obtains on average over 98\% accuracy.

In addition to the CLIP detection model, we evaluated the methods of \cite{cnndet} and \cite{dire} on the test set. In particular, we initialized the ResNet architecture of CNNDet with the authors' weights; this is obtained from training with a data augmentation of scheme of randomly blurring or JPEG compressing an image with probability 0.5 \cite{cnndet}.

For the DIRE method \cite{dire}, we initialize the diffusion model with \(256 \times 256\) unconditional weights obtained from \cite{ADM}. The binary classifier to distinguish real or fake images given the DIRE representation is also based on a ResNet architecture. We initialize the classifier with weights from \cite{dire}. In the second case, we refer to the method of computing an image's DIRE representation and feeding it to the classifier simply as \textit{DIRE}.

One observation that must be addressed is the very poor performance of DIRE on the real images in our test set. To confirm that there were no bugs in our evaluation of DIRE, we replaced the real images in our test set with 1000 real images from the test set given by the authors of \cite{dire}. As Table \ref{tab:realcomp} shows, on these images, we were able to reproduce the results of \cite{dire}. In addition, the CLIP model still performed well on the real images from \cite{dire}. 

\begin{table}[htbp]
\centering
\caption{Accuracy of CLIP and DIRE on different sets of real images}
\begin{tabular}{|l|c|c|}
\hline
 & Our real images & \cite{dire}'s real images \\
 \hline
 CLIP & .957 & .908 \\
 DIRE & .002 & .988 \\
 \hline
\end{tabular}
\label{tab:realcomp}
\end{table}

Next, observe that CNNDet performs very well on ProGAN generated images, which is sensible since ProGAN was the sole generation method for the training data in \cite{cnndet}. CNNDet shows some generalization capability for images generated by StyleGAN and Diff-StyleGAN2, but does quite poorly on diffusion generated images. We expect that with finetuning the performance of CNNDet would increase as was shown in \cite{towards}. 

Finally, if we ignore the performance of DIRE on the real images in our test set, we observe that DIRE and CLIP obtain similar results. Considerations of speed and cost (in terms of GPU RAM) would lead one to prefer CLIP over DIRE for AIGI detection. The costly aspect of DIRE is the requirement of a diffusion model to compute DIRE representations of images. It took several hours on a NVIDIA GTX 4090 GPU with 24 GB VRAM to compute the DIRE representations of the 11,000 images in our test set. For comparison, results from CLIP were obtained in less than 10 minutes on a laptop with a NVIDIA RTX 3050 Ti GPU with 4 GB VRAM.

\begin{table}[htbp]
\centering
\caption{CLIP, DIRE, and CNNDet accuracy by generation method}
\begin{tabular}{|l|c|c|c|}
\hline
Generation Method & 	CNNDET & 	DIRE & 	CLIP \\
\hline
ADM & .003 & \textbf{1.0} & .993 \\
DDPM & .005 & \textbf{.997} & \textbf{.997} \\
Diff-ProjectedGAN & .045  & .999 & \textbf{1.0} \\
Diff-StyleGAN2 & .804 & \textbf{1.0} & \textbf{1.0} \\
IDDPM & .004 & \textbf{1.0} & \textbf{1.0} \\
LDM & .004 & \textbf{1.0} & .999 \\
PNDM & .002 & \textbf{1.0} & \textbf{1.0} \\
ProGAN & .998  & .999 & \textbf{1.0} \\
ProjectedGAN & .074 & \textbf{1.0} & \textbf{1.0} \\
Real & \textbf{1.0} & .002 & .957 \\
StyleGAN & .319 & .998 & \textbf{.999} \\
\hline
\end{tabular}
\label{tab:acc}
\end{table}

\begin{table}[htbp]
\centering
\caption{Number of correct predictions made by CLIP, DIRE, and CNNDet}
\begin{tabular}{|l|c|c|c|c|}
\hline
Generation Method & 	Total Images & 	CNNDET & 	DIRE & 	CLIP \\
\hline
ADM & 978 & 3 & \textbf{978} & 971 \\
DDPM & 1037 & 5 & \textbf{1034} & \textbf{1034} \\
Diff-ProjectedGAN & 977 & 44 & 976 & \textbf{977} \\
Diff-StyleGAN2 & 1009 & 811 & \textbf{1009} & \textbf{1009} \\
IDDPM & 986 & 4 & \textbf{986} & \textbf{986} \\
LDM & 962 & 4 & \textbf{962} & 961 \\
PNDM & 1010 & 2 & \textbf{1010} & \textbf{1010} \\
ProGAN & 993 & 991 & 992 & \textbf{993} \\
ProjectedGAN & 1000 & 74 & \textbf{1000} & \textbf{1000} \\
Real & 1000 & \textbf{1000} & 2 & 957 \\
StyleGAN & 1048 & 344 & 1046 & \textbf{1047} \\
\hline
\end{tabular}
\label{tab:count}
\end{table}

\begin{table*}[htbp]
\centering
\caption{Confusion matrix and metrics for CLIP. Rows correspond to ground truth labels and columns correspond to predicted labels.}
\scalebox{0.7}{

\begin{tabular}{|l|c|c|c|c|c|c|c|c|c|c|c|c|c|c|}
\hline
 & 	ADM & 	DDPM & 	DPjG & 	DSG & 	IDDPM & 	LDM & 	PjG & 	SG & 	PG & 	PNDM & 	Real & 	Class Precision & 	Class Recall & 	Class F1 \\
\hline
ADM & 718 & 2 & 0 & 0 & 251 & 0 & 0 & 0 & 0 & 0 & 7 & .87 & .734 & .796 \\
DDPM & 0 & 1032 & 2 & 0 & 0 & 0 & 0 & 0 & 0 & 0 & 3 & .968 & .995 & .981  \\
DPjG & 0 & 0 & 821 & 0 & 0 & 2 & 151 & 0 & 3 & 0 & 0 & .775 & .84 & .806 \\
DSG & 0 & 0 & 0 & 1008 & 0 & 0 & 0 & 0 & 1 & 0 & 0 & .994 & .999 & .997 \\
IDDPM & 94 & 7 & 0 & 0 & 884 & 0 & 1 & 0 & 0 & 0 & 0 & .776 & .897 & .832 \\
LDM & 0 & 0 & 2 & 0 & 0 & 953 & 0 & 0 & 5 & 1 & 1 & .996 & .991 & .993 \\
PjG & 0 & 0 & 227 & 1 & 0 & 0 & 766 & 1 & 5 & 0 & 0 & .831 & .766 & .797 \\
SG & 0 & 0 & 1 & 3 & 0 & 0 & 1 & 1034 & 8 & 0 & 1 & .994 & .987 & .99 \\
PG & 0 & 0 & 4 & 1 & 0 & 0 & 3 & 5 & 980 & 0 & 0 & .978 & .987 & .982 \\
PNDM & 0 & 0 & 1 & 1 & 0 & 2 & 0 & 0 & 0 & 1006 & 0 & .999 & .996 & .998 \\
Real & 13 & 25 & 1 & 0 & 4 & 0 & 0 & 0 & 0 & 0 & 957 & .988 & .957 & .972 \\
\hline
\end{tabular}
}

\label{tab:clip}
\end{table*}

\section{Conclusion}\label{section_conclusion}
Detection of AI-generated images (AIGI) is both an interesting academic problem and a vital task for the internet at large, as misuse of AIGI can poison training datasets, ignite copyright disputes, and produce disinformation on a massive scale. In this paper, we fine-tuned a pre-trained CLIP model on pairs of AIGI and text strings corresponding to the generation source, as well as on real photographic images. We compared our CLIP model to two state-of-the art models with custom architectures designed to detect AIGI, CNNDet and DIRE. Evaluation of these models included the task of differentiating AIGI from real images, as well as classifying each image by its generation source model. 

We find that our CLIP model performed well, with an accuracy greater than 90\%, on GAN-generated images, diffusion-generated images, and real photographs. In contrast, while DIRE performed well on AIGI, it struggled on our dataset of real images. CNNDet handled real/AIGI determination well but struggled to identify the generation source of AIGI.  

Our results have wider implications for both detection of AIGI as well as the role of internet-scale pre-trained models in computer vision. We show how  a general architecture combined with massive pre-training datasets can match or surpass models whose architecture is custom built for computer vision tasks. These specialized models can have issues adapting to new or wider datasets; pre-training datasets may be diverse enough that new data does not impact a pre-trained model as much. CLIP was even able to identify the generation source between AIGI generated by models based on its massive pre-training.

These results can improve the accessibility of tools for detecting AIGI, which will increase the ability of non-technical organizations to handle growing AIGI problems. Rather than relying on specialized architectures which may require technical knowledge to retrain or deploy, users can implement up-to-date pre-trained general models and replace fine tuned weights over time. We also see that our model requires much less VRAM and time to run than custom models such as DIRE, allowing a wider variety of users to deploy the model on inexpensive, commercially available GPUs. Finally, these results imply that massive pretrained models, such as multi-modal Large Language Models, may be able to take on complex computer vision tasks or pick up on subtle image source details regardless of image content.   

\backmatter

\bmhead{Acknowledgements}
No funds, grants, or other support was received for this work.

\bmhead{Data Availability}
 No new data was generated during the production of this work. The data used in this work can be found at \textit{zenodo.org/records/7528113} and \textit{github.com/fyu/lsun.} The specific CLIP implementation used in this work can be found at \textit{github.com/openai/CLIP}.


\bibliography{sn-bibliography}

\end{document}